\documentclass[10pt,twocolumn,letterpaper]{article}

\usepackage{cvpr}
\usepackage{times}
\usepackage{epsfig}
\usepackage{graphicx}
\usepackage{amsmath}
\usepackage{amssymb}
\usepackage{booktabs}
\usepackage{mathptmx}
\usepackage[table,xcdraw]{xcolor}

\usepackage[pagebackref=true,breaklinks=true,letterpaper=true,colorlinks,bookmarks=false]{hyperref}

\cvprfinalcopy 

\def\cvprPaperID{26} 

\ifcvprfinal\fi
\begin{document}

\title{Efficient 2.5D Hand Pose Estimation via Auxiliary Multi-Task Training for Embedded Devices}

\author{Prajwal Chidananda,  Ayan Sinha,   Adithya Rao,  Douglas Lee, and Andrew Rabinovich\\ 
Magic Leap, Inc.\\
{\tt\small (pchidananda, asinha, arao, dlee, arabinovich) @magicleap.com}
}

\maketitle

\begin{abstract}
   2D Key-point estimation is an important precursor to 3D pose estimation problems for human body and hands. In this work, we discuss the data, architecture, and training procedure necessary to deploy extremely efficient 2.5D hand pose estimation on embedded devices with highly constrained memory and compute envelope, such as AR/VR wearables. Our 2.5D hand pose estimation consists of 2D key-point estimation of joint positions on an egocentric image, captured by a depth sensor, and lifted to 2.5D using the corresponding depth values. Our contributions are two fold: (a) We discuss data labeling and augmentation strategies, the modules in the network architecture that collectively lead to $3\%$ the flop count and $2\%$ the number of parameters when compared to the state of the art MobileNetV2
   \cite{sandler2018mobilenetv2} architecture. (b) We propose an auxiliary multi-task training strategy needed to compensate for the small capacity of the network while achieving comparable performance to MobileNetV2. Our 32-bit trained model has a memory footprint of less than 300 Kilobytes, operates at more than 50 Hz with less than 35 MFLOPs. 
\end{abstract}

\section{Introduction}

Hand pose estimation is a critical component of mixed reality (MR) applications to enable controller-less interactions. It comes in different forms.

\begin{itemize}
	\item Discrete pose classification: The objective is to classify the hand pose into one of predefined classes such as OK, thumbs up, open hand etc. \cite{isaacs2004hand}
	\item 2D hand key-point estimation: Here, a select number of key-points on the hand are detected in the image. These usually correspond to visible joint positions of the hand skeleton \cite{simon2017hand}. 
	\item 2.5D hand key-point estimation: Similar to 2D hand key-point estimation, select skeletal joint positions are detected in the image. Additionally, the key-points are lifted to 2.5d using corresponding depth values of the estimated 2D joint pixels in the image. 
	\item 3D hand key-point estimation: All key-points corresponding to skeletal joint positions are detected in 3D coordinates. The 3D detection corresponds to plausible anthropomorphic configurations of the human hand \cite{moon2018v2v, mueller2018ganerated}. 
	\item Fully articulated 3D hand: The full mesh of the hand is detected and tracked. This corresponds to 3D hand key-point estimation in conjunction with 3D hand shape estimation \cite{pavlakos2019expressive, xiang2018monocular}.
\end{itemize}

In this article we focus on 2.5D hand key-point estimation. While such an approach suffices for most hand interactions in AR/VR/MR environments, it fails when the key-points on the hand are self-occluded and the depth corresponds to hand surface's depth ignoring occlusions. 

\section{Data, labeling, and augmentation}
The input to the vision-based hand tracking systems is often either a monocular rgb/grayscale image or a depth image. Depth-based approaches have been shown to outperform RGB-based approaches for 3D pose estimation \cite{baek2018augmented, choi2017learning, choi2017robust, mueller2017real, yuan2018depth, sinha2016deephand}. For our implementation, we use a time-of-flight (TOF) depth sensor.  

Using mechanical turk, we labeled $17$ key-points per image. The labelers only label the visible key-points which correspond to all. Although the network only predicts 8 key-points, the additional key points serve as auxiliary supervision (described in section \ref{training}). Images of hands are also labeled with 8 discrete hand key-pose classes, as well as right/left hand assignments. Later, we describe how additional labels act as supervisory tasks.

To avoid hand-like (distractor) objects confounding the predictions, we composite the images containing the hand with varied backgrounds containing challenging distractor objects. By collecting data in controlled environments and using augmentation, we expand the training data to generalize to different environments. As most of the collected data is for a user performing single handed interactions, a skew is introduced in the dataset.
To mitigate this, left and right hands are composited from different images to look like two handed interactions. 

\begin{figure*}[ht]
 \begin{center}
        \includegraphics[width=\linewidth]{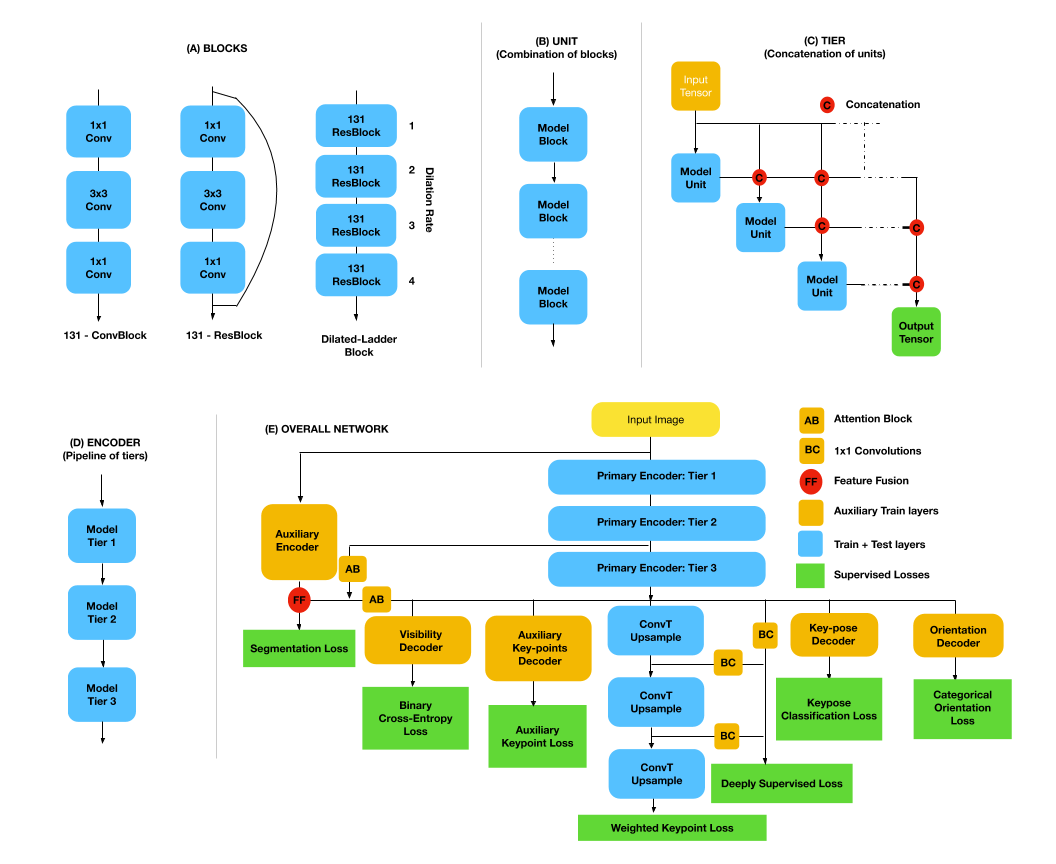}
 \caption{(A) Model Blocks used in our network architecture (B) Model Units: a combination of model blocks, (C) Model Tier: a concatenation of model units, which could be seen as a variant of the DenseBlock, (D) Primary encoder: a pipeline of three tiers, (E) Overall network architecture along with auxiliary modules with corresponding supervised losses trained end-to-end for our 2.5D hand key-point estimation. Our Tier-1 is a simple convolution layer, Tier-2 is composed of two 131-ConvBlocks, and Tier-3 is composed of two Dilated-Ladder units.}
 \label{fig:arch}
 \end{center}
\end{figure*}

\section{Network Modeling}
TOF depth sensors capture phase images which are then translated to a depth image using post-processing. The compute time for post-processing the phase images to calculate the depth image adds a considerable portion to the end-to-end latency for hand tracking. Towards addressing this issue, we explore directly using a linear combination of phase images, which we term \textit{amplitude} image to perform 2D key-point estimation and perform depth image processing in parallel, this effectively reducing the overall latency. As discussed later, this improves performance while removing around 20ms latency of the sequential depth processing by doing it in parallel. 

Our network follows an encoder-decoder structure. The encoder is is organized as a pipeline of tiers; where each tier is a concatenation of units and each unit is a combination of blocks as shown in Fig. \ref{fig:arch}.
We use a variant of the DenseNet \cite{huang2017densely} block in Tier-2, wherein we use two Conv-BN-ReLU model blocks instead of the conventional four blocks in order to reduce sequential compute. Unlike in DenseNet, we do not concatenate the input tensor with the intermediate feature tensors. This reduces compute and stems from our observation that the input feature tensor is mostly redundant and does not contribute to performance gains. However, removing the input tensor slows down training. Hence, in Tier-3 we replace the standard Conv-BN-ReLU model blocks with ResNet blocks to improve information flow during back-propagation. Furthermore, we use a ladder of bottleneck ResNet blocks wherein each module has convolution with a different dilation rate, i.e., 4 bottleneck ResNet blocks with a dilation rate of 1,2,3,4 respectively. Overall, our network combines the best of DenseNet's feature reuse property and ResNet's \cite{he2016deep} information flow property. 

We also use small number of channels to reduce parallel compute with 16 channels after Tier-1, 32 channels after Tier-2, and 64 channels after Tier-3. Furthermore, we use dilated convolutions in Tier-3 to increase the receptive field of our relatively shallow network. 


We employ grouped convolutions at all but first tier so as to reduce compute. In Tier-2 we use grouping factor of 4 and we use grouping factor of 8 in Tier-3. In our experiments, we observed that the encoder is responsible for a majority of the gains in performance, and changing the decoder architecture only marginally affects performance. Hence, our key-point decoder is very lean and all convolutions are channel wise, i.e., number of groups is equal to the number of channels. 

\subsection*{Embedded Implementation}
The embedded implementation of our network architecture is carefully designed to reduce compute / memory overhead and energy consumption. The Myriad2 VPU runs at 600MHz and provides 12 VLIW compute cores called SHVs. Under typical workloads, total power consumption of the Myriad2 chip is less than 2W. Each SHV is allocated a 128KB working memory slice out of 2MB on-chip SRAM. Our Gestures DNN must perform real-time 45FPS hand tracking and gesture recognition using a single SHV.

Using conventional implementations makes these minimization challenging for several reasons: (a) Typical deep learning frameworks convert convolution into a large matrix multiply in a highly memory inefficient way (b) Input data tensors are typically stored in channel planar format, which is inefficient for vectorization. (c) The kernel sizes are often not a multiple of the vector instruction width (d) Off-the-shelf implementations of dilated convolutions have considerable computational overhead.

To address these challenges, (a) We reorder and interleave the input data channels to each convolution layer to align with vector instruction ordering. We simultaneously reorder the kernel stack such that convolutions are reduced to dot products and the output channel is automatically encoded in the interleaved format for the next layer. (b) We group convolution layers so that the number of filters per group is a multiple of the register lanes, consequently, maximizing vector register utilization. (c) We use a comb design for dilated convolutions which minimizes the on-chip memory footprint. For example, for dilation=2, convolution splits into four independent field-wise computations (even rows-even columns, even-odd, etc) which are computed independently and recombined on output. Dilated convolutions are thus computed at zero effective overhead.

\section{Training}\label{training}
We adopt a multi-task learning paradigm in the training procedure by employing multiple network predictions and loss functions, while maintaining the the prime focus on 2D key-point prediction. Note that human labeling for auxiliary tasks such as discrete hand pose or categorical orientation is much faster than key-point labeling, or comes for free with key-point labeling such as key-point visibility. At inference time, only the primary encoder and decoder are part of the network running on device.

Cross entropy with 1-hot label is used to predict each of the 16 key-points (8 key-points per hand). We observed that the aggressive down-sampling in the early layers coupled with the low network capacity, makes the conventional mean squared loss (MSE) loss ineffective. Cross entropy has a stronger gradient signal and is much more effective. We experimented with label smoothing but did not observe performance gains. 




Given that we have 17 key-point labels per hand, we use the additional key-points as training supervision, even though they are not part of the final inference module. As our decoder is composed of fully grouped convolutions, we observed some key-points failing to train all together. Hence, the decoder for the auxiliary key-points  do not use grouping in the convolutions so as to avoid floating key-points and regularize the feature tensor after the encoder. 

The binary key-point and hand visibility masks serve three purposes: make training stable, suppress occluded key-points and invisible hands during inference, enable an early out during inference to reduce latency. We use a binary cross entropy loss to train these tasks.


The data collected was heavily skewed against palm-facing data causing the trained model to under-perform on palm-facing data. To address this, we regularized the predictions using a categorical hand orientation loss. We label $8$ categorical hand orientations that could arise from supination or pronation of the hand. Since the categories are approximate, we soften the labels and use cross-entropy loss to train these tasks.


We classify the hand pose into nine discrete classes: OK, open-hand, pinch, C-pose, fist, L-pose, point, thumbs-up, and a dummy class capturing all other poses. We use cross-entropy loss to train the discrete hand-pose classification. 

Following the architecture of BiseNet proposed \cite{yu2018bisenet}, we use our network architecture discussed previously as the context path and use a spatial path similar to BiseNet as a training-time artifact to provide useful supervision that regularizes the key-points that jump off the hand. We train the network to segment three classes: background, left hand and right hand using a per-pixel cross entropy loss.

We observe that key-points often fail to train due to the grouped structure of decoder. Following \cite{wang2015training}, we add additional key-point supervision heads after three intermediate layers, with different spatial resolutions: tier 3 of the primary encoder ($1/8\emph{th}$ the full resolution), the first up-sampling block ($1/4\emph{th}$ the full resolution) and the second up-sampling block (1/2 the full resolution). This stabilizes training and facilitates better gradient flow for training.

Our final loss is a weighted sum of all the individual task losses: primary key-point loss $\mathcal{L}_{kp}$, auxiliary key-point loss $\mathcal{L}_{akp}$, key-point and hand visibility loss $\mathcal{L}_{kphv}$, categorical hand orientation loss $\mathcal{L}_{cho}$, discrete hand pose loss $\mathcal{L}_{dhp}$, segmentation loss $\mathcal{L}_{seg}$, deep supervision loss $\mathcal{L}_{ds}$.

We use task-weighting to weigh the different losses, as the  predictions are not all at the same scale. The weights for the different tasks were derived heuristically, but can be replaced with an adaptive weighting using gradient normalization as described in \cite{chen2017gradnorm}. The full training loss with all task losses and corresponding weights are shown in Eq \ref{full_loss}.

\begin{multline}
        \mathcal{L}_{total} = w_{kp}\mathcal{L}_{kp} + w_{akp}\mathcal{L}_{akp} + w_{kphv}\mathcal{L}_{kphv} + \\
        w_{cho}\mathcal{L}_{cho} + w_{dhp}\mathcal{L}_{dhp} + w_{seg}\mathcal{L}_{seg} + w_{ds}\mathcal{L}_{ds}
    \label{full_loss}
\end{multline}
where $w_{kp}=1$, $w_{akp}=1$, $w_{kphv}=20$, $w_{cho}=20$, $w_{dhp}=10$, $w_{seg}=50$, $w_{ds}=1$.

We empirically observe that the network finds it harder to predict finger tips when compared to the other key-points. We address this by simply doubling the losses for finger tips while calculating  $\mathcal{L}_{kp}$ and $\mathcal{L}_{akp}$.



\begin{table}[ht]
\centering
\caption{Ablation study: Each supervision technique is removed and the corresponding average key-point errors are shown. $\mathcal{W}_{k}$ here represents key-point loss weighting.}
\resizebox{\columnwidth}{!}{%
\begin{tabular}{@{}lccccccc@{}}
\toprule
Input mode & \textbf{$\mathcal{L}_{seg}$} & \textbf{$\mathcal{L}_{ds}$} & \textbf{$\mathcal{L}_{cho}$} & \textbf{$\mathcal{W}_{k}$} & {$\mathcal{L}_{kphv}$} & \textbf{$\mathcal{L}_{dhp}$} & Err (px)  \\ \midrule
Depth               & $\times$              & $\times$                  & $\times$                         & $\times$                    & $\times$                       & $\times$          & 6.226                        \\
Amplitude           & $\times$              & $\times$                  & $\times$                         & $\times$                    & $\times$                       & $\times$          & 6.054                        \\
Amplitude           & $\checkmark$          & $\checkmark$              & $\times$                         & $\checkmark$                & $\checkmark$                   & $\checkmark$      & 5.941                        \\
Amplitude           & $\checkmark$          & $\checkmark$              & $\checkmark$                     & $\times$                    & $\checkmark$                   & $\checkmark$      & 5.730                        \\
Amplitude           & $\checkmark$          & $\times$                  & $\checkmark$                     & $\checkmark$                & $\checkmark$                   & $\checkmark$      & 5.994                        \\
Amplitude           & $\times$              & $\checkmark$              & $\checkmark$                     & $\checkmark$                & $\checkmark$                   & $\checkmark$      & 5.690                        \\
Amplitude           & $\checkmark$          & $\checkmark$              & $\checkmark$                     & $\checkmark$                & $\checkmark$                   & $\times$          & 5.655                        \\

Depth               & $\checkmark$          & $\checkmark$              & $\checkmark$                     & $\checkmark$                & $\checkmark$                   & $\checkmark$      & 5.898                        \\
Amplitude           & $\checkmark$          & $\checkmark$              & $\checkmark$                     & $\checkmark$                & $\checkmark$                   & $\checkmark$      & $\mathbf{5.556}$                        \\ \bottomrule
\end{tabular}%
}
\label{tab:my-table}
\end{table}
\begin{table}[ht]
\centering
\caption{Comparison of size, computational cost and performance between DenseNet, MobileNet-V2 and Our implementation.}
\resizebox{\columnwidth}{!}{%
\begin{tabular}{@{}lllc@{}}
\toprule
{Backbone} & {Parameters} & {GFLOPs} & {Keypoint Error} \\ \midrule
DenseNet          & 7.017M                    & 4.866                     & $\mathbf{4.457 px}$                  \\
MobileNetV2       & 1.893M                    & 1.209                    & 5.306 px                       \\
Ours              & $\mathbf{0.041M}$                    &  $\mathbf{0.035}$               & 5.556 px                        \\ \bottomrule
\end{tabular}%
}
\label{tab:compare_dense_mobile}
\end{table}

\begin{table}[ht]
\centering
\caption{Comparison of inference time for a single forward pass on an Intel Movidius Myriad between mvTensor and our custom implementation.}
\label{tab:inference-table}
\begin{tabular}{@{}llllll@{}}
\toprule
Shave \#      & 1      & 2                        & 3                        & 4                        & 6                        \\ \midrule
mvTensor (ms) & 205.0 & 114.0                   & 84.0                    & 68.0                    & 54.0                    \\
Ours (ms)     & $\mathbf{16.0}$     & - & - & - & - \\ \bottomrule
\end{tabular}%
\end{table}

\section{Results and Discussion}

\begin{figure*}[ht]
 \begin{center}
        \includegraphics[width=\linewidth]{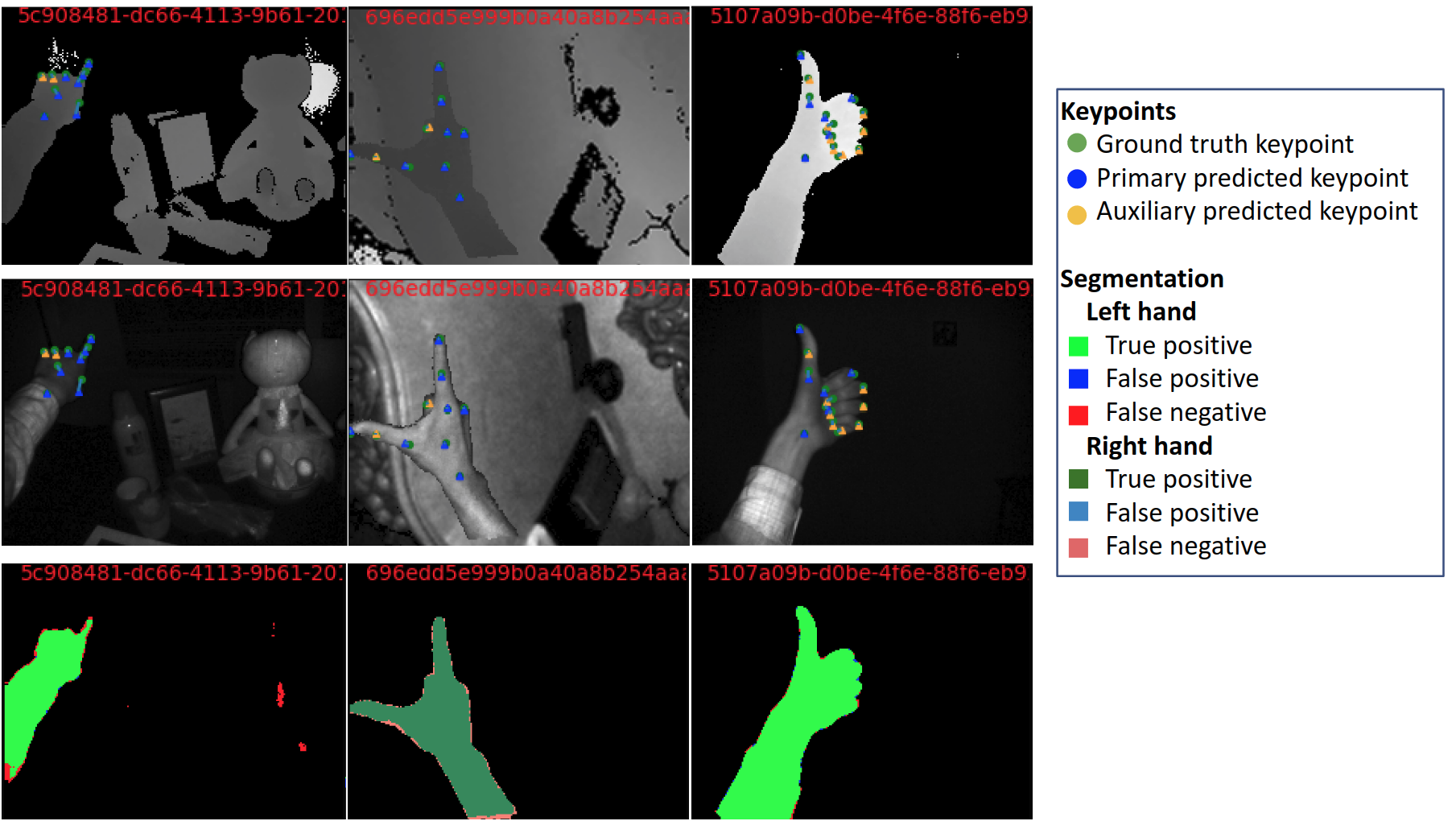}
 \caption{Qualitative results on our in-house dataset. Top row shows 2D key point predictions on depth images, Middle row shows 2D key point predictions on amplitude images, Bottom row shows corresponding segmentation results.}
 \label{fig:qual}
 \end{center}
\end{figure*}

We evaluate our method on its 2D key-point estimation performance and run time performance on an Intel Movidius Myriad chip, which affects latency. We run experiments on a reduced in-house dataset that has 40 users using a train/validation/test split of 28/4/8. The dataset has over 5K frames per user, with a total 225733 frames. To quantitatively evaluate the key-point estimation, we report the average 2D euclidean distance between the estimated 2D key-points and the ground truth, expressed in pixels. 

First, we evaluate the impact of all the losses used in the training scheme by doing an ablation study. The models are all trained for 20 epochs, with a learning rate of 0.001 with a linear decay, using Adam as the optimizer. As shown in Table \ref{tab:my-table}, each loss positively contributed to the performance and the model which has all the training losses performs best. Furthermore, the amplitude image as input to the network consistently performs better than the depth image input, while also reducing latency. Note, at run-time we lift the 2D predictions on amplitude image to 2.5D using the depth image processed in parallel. We use simple filtering and heuristics to ensure valid depth for identified 2D key-points.   

Second, we compare our model with DenseNet-121 \cite{huang2017densely} and MobileNet-V2 \cite{sandler2018mobilenetv2} in terms of the number of model parameters, the total number of floating point operations for a single frame, and the 2D key-point error. We use the same decoder for all three networks, with the same training scheme used in the ablation study. Table \ref{tab:compare_dense_mobile} compares the three models. We see that our model is $2\%$ the size of MobileNetV2 in terms of parameters and has $3\%$ the compute of MobileNetV2, with only 4.7\% degradation in key-point localization performance.  

We evaluate the run time performance of our implementation by comparing its latency with that of the off-the-shelf mvTensor implementation on the Intel Movidius Myriad \cite{ionica2015movidius}. Table \ref{tab:inference-table} compares the two implementations. By reordering of input activation channels and the kernel stack, having grouped convolution layers with a multiple-of-the-register-length number of kernels per group, and using an efficient comb design for dilation, our embedded implementation achieves 16ms latency, which is a 12x speedup over mvTensor. 


Overall, we describe an end-to-end 2.5D key-point estimation system with unique insights at each stage of the pipeline: data, network architecture, training and embedded implementation. The first take-away is that auxiliary labeling and multi-task training is a viable strategy to improve performance of extremely compact network architectures bound by compute constraints on edge devices. Second, network architectures crafted in sync with embedded hardware operating constraints can offer significant benefits at run-time, vital for latency critical applications like hand-tracking.







\section{Acknowledgements} We would like to acknowledge Lexin Tang at Magic Leap, Inc. for her work on the embedded code implementation.

{\small
\bibliographystyle{ieee_fullname}
\bibliography{egbib}
}

\end{document}